\documentclass[10pt,twocolumn,letterpaper]{article}
\usepackage{cvpr}
\usepackage{times}
\usepackage{epsfig}
\usepackage{graphicx}
\usepackage{amsmath}
\usepackage{amssymb}
\usepackage{gensymb} %
\usepackage{siunitx}
\usepackage{float}
\usepackage{booktabs}
\usepackage{colortbl}
\usepackage{xcolor}
\usepackage{multirow}
\usepackage{makecell}
\usepackage{wrapfig}
\usepackage[export]{adjustbox}
\usepackage{subcaption}
\usepackage{tabularx}
\usepackage[T1]{fontenc}
\usepackage[labelsep=period]{caption}
\RequirePackage{enumitem}
\setlist[itemize]{nosep}
\newcommand{\Tref}[1]{Table~\ref{#1}}

\newcommand{\fref}[1]{Fig.~\ref{#1}}
\newcommand{\Fref}[1]{Figure~\ref{#1}}

\usepackage{xcolor}
\newcounter{todos}
\AtEndDocument{\ifnum\value{todos}>0 \PackageWarning{TODOS}{There are \arabic{todos} todos left in this paper! Fix them before submitting the paper!} \fi}

\renewcommand{\paragraph}[1]{\vspace{0.2em}\noindent \textbf{#1 \hspace{0.2em}}}

\usepackage[pagebackref=true,breaklinks=true,letterpaper=true,colorlinks,bookmarks=false]{hyperref}

\cvprfinalcopy  %

\def\cvprPaperID{2939} %

\newcommand{\B}{\textbf}
\newcommand{\Frst}[1]{\textcolor{red}{\textbf{#1}}}
\newcommand{\Scnd}[1]{\textcolor{blue}{\textbf{#1}}}
\newcommand{\scene}[1]{{\sc #1}\xspace}

\newcommand{\RefineStage}{RefineNet\xspace}
\newcommand{\CoarseStage}{CoarseNet\xspace}
\newcommand{\CoarseStagenospace}{CoarseNet}

\newcommand{\weightnetwork}{weight network\xspace}

\newcommand{\flownetwork}{flow network\xspace}
\newcommand{\featurealignmodule}{deformable alignment module\xspace}
\newcommand{\featurefusionmodule}{temporal attention fusion module\xspace}
\newcommand{\expolowthr}{0.15}
\newcommand{\expohighthr}{0.9}
\newcommand{\SingleStageRefineStage}{RefineNet$^\dag$\xspace}

\newcommand{\twoexposure}{two-exposure\xspace}
\newcommand{\threeexposure}{three-exposure\xspace}
\newcommand{\ntwoExposure}{2-Exposure\xspace}
\newcommand{\nthreeExposure}{3-Exposure\xspace}
\newcommand{\lowexposure}{low-exposure\xspace}

\newcommand{\highexposure}{high-exposure\xspace}
\newcommand{\LowExposure}{Low-Exposure\xspace}
\newcommand{\MiddleExposure}{Middle-Exposure\xspace}
\newcommand{\HighExposure}{High-Exposure\xspace}
\newcommand{\AllExposure}{All-Exposure\xspace}

\newcommand{\PSNRT}{PSNR\xspace}

\newcommand{\VQM}{HDR-VQM\xspace}
\newcommand{\VDP}{HDR-VDP2\xspace}

\newcommand{\crf}{\mathcal{F}}
\newcommand{\hdr}{H}
\newcommand{\tmhdr}{T}
\newcommand{\gttmhdr}{\tilde{\tmhdr}}
\newcommand{\ldr}{L}

\newcommand{\originldr}{\tilde{\ldr}}

\newcommand{\expt}{t}

\newcommand{\coarsehdr}{\hdr^c}
\newcommand{\coarsetmhdr}{\tmhdr^c}
\newcommand{\coarseloss}{\mathcal{L}^\text{c}}
\newcommand{\refinehdr}{\hdr^r}

\newcommand{\finalhdr}{\hdr}
\newcommand{\finaltmhdr}{\tmhdr}
\newcommand{\refineloss}{\mathcal{L}^r}

\newcommand{\validmask}{M}
\newcommand{\feature}{F}
\newcommand{\alignedfeature}{\tilde{F}}

\newcommand{\staticdatalong}{\emph{static scenes with GT}\xspace}
\newcommand{\dynamicgtdatalong}{\emph{dynamic scenes with GT}\xspace}
\newcommand{\dynamicdatalong}{\emph{dynamic scenes without GT}\xspace}
\newcommand{\staticdata}{$\mathcal{D}_s^{gt}$\xspace}
\newcommand{\dynamicgtdata}{$\mathcal{D}_d^{gt}$\xspace}
\newcommand{\dynamicdata}{$\mathcal{D}_d$\xspace}
\newcommand{\kalantaridata}{\emph{Kalantari13} dataset\xspace}

\newcommand{\NumOfStaticTwoExp}{49}
\newcommand{\NumOfStaticThreeExp}{48}
\newcommand{\NumOfDynamicGTTwoExp}{76}
\newcommand{\NumOfDynamicGTThreeExp}{108}
\newcommand{\NumOfDynamicTwoExp}{50} %
\newcommand{\NumOfDynamicThreeExp}{50} %

\newcommand{\pokerscene}{\scene{Poker Fullshot}}
\newcommand{\carouselscene}{\scene{Carousel Fireworks}}
\newcommand{\ThrowTowelScene}{\scene{Throwing Towel 2Exp}}

\newcommand{\YanCVPR}{Yan19~\cite{yan2019attention}\xspace}
\newcommand{\KalantariTOG}{Kalantari13~\cite{kalantari2013patch}\xspace}
\newcommand{\KalantariEG}{Kalantari19~\cite{kalantari2019deep}\xspace}

\ifcvprfinal\fi
\begin{document}

\title{HDR Video Reconstruction: A Coarse-to-fine Network and A Real-world Benchmark Dataset}

\author{Guanying Chen$^{1,2}$ \quad Chaofeng Chen$^{1}$ \quad Shi Guo$^{2,3}$ \quad Zhetong Liang$^{2,3}$ \\ \quad Kwan-Yee K. Wong$^1$ \quad Lei Zhang$^{2,3}$ \vspace{0.3em} \\
{\normalsize $^1$Department of Computer Science, The University of Hong Kong} \quad
{\normalsize $^2$DAMO Academy, Alibaba Group} \quad \\
{\normalsize $^3$Department of Computing, The Hong Kong Polytechnic University}
}

\maketitle

\begin{abstract}
    High dynamic range (HDR) video reconstruction from sequences captured with alternating exposures is a very challenging problem. Existing methods often align low dynamic range (LDR) input sequence in the image space using optical flow, and then merge the aligned images to produce HDR output. However, accurate alignment and fusion in the image space are difficult due to the missing details in the over-exposed regions and noise in the under-exposed regions, resulting in unpleasing ghosting artifacts. To enable more accurate alignment and HDR fusion, we introduce a coarse-to-fine deep learning framework for HDR video reconstruction. Firstly, we perform coarse alignment and pixel blending in the image space to estimate the coarse HDR video. Secondly, we conduct more sophisticated alignment and temporal fusion in the feature space of the coarse HDR video to produce better reconstruction. Considering the fact that there is no publicly available dataset for quantitative and comprehensive evaluation of HDR video reconstruction methods, we collect such a benchmark dataset, which contains $97$ sequences of static scenes and $184$ testing pairs of dynamic scenes. Extensive experiments show that our method outperforms previous state-of-the-art methods. Our code and dataset can be found at \url{https://guanyingc.github.io/DeepHDRVideo}.

\end{abstract}

\section{Introduction}
\begin{figure}[t] \centering
    \input{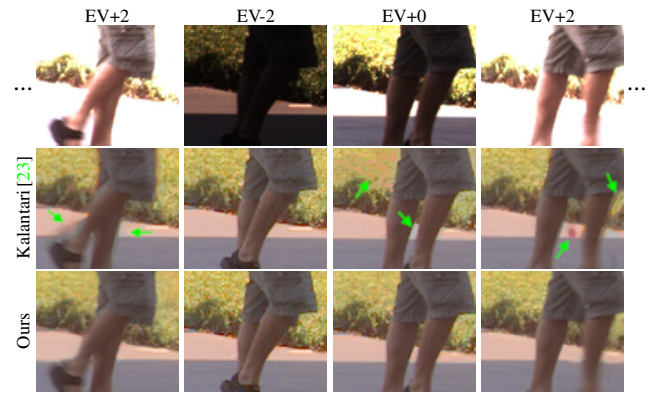} 
    \caption{HDR video reconstruction from sequences captured with three alternating exposures. Row~1 shows four input LDR frames. Rows 2--3 are the reconstructed (tonemapped) HDR frames.} \label{fig:teaser}
\end{figure}

Compared with low dynamic range (LDR) images, high dynamic range (HDR) images can better reflect the visual details of a scene in both bright and dark regions.
Although significant progress has been made in HDR image reconstruction using multi-exposure images~\cite{kalantari2017deep,wu2018deep,yan2019attention},
the more challenging problem of HDR video reconstruction is still less explored. 
Different from HDR image reconstruction, HDR video reconstruction has to recover the HDR for every input frame (see \fref{fig:teaser}), but not just for a single reference frame (\eg, the middle exposure image).
Existing successful HDR video reconstruction techniques often rely on costly and specialized hardware (\eg, scanline exposure/ISO or internal/external beam splitter)~\cite{tocci2011versatile,kronander2014unified,zhao2015unbounded}, which hinders their wider applications among ordinary consumers.
A promising direction for low-cost HDR video reconstruction is to utilize video sequences captured with alternating exposures (\eg, videos with a periodic exposure of \{EV-3, EV+3, EV-3, $\dots$\}).
This is practical as many off-the-shelf cameras can alternate exposures during recording.

Conventional reconstruction pipeline along this direction often consists of two steps~\cite{kang2003high}. 
In the first step, neighboring frames with different exposures are aligned to the current frame using optical flow. 
In the second step, the aligned images are fused to produce the HDR image.
However, accurate alignment and fusion are difficult to achieve for LDR images with different exposures as there are saturated pixel values in the over-exposed regions, and noise in the under-exposed regions.
Recently, Kalantari and Ramamoorthi~\cite{kalantari2019deep} proposed to estimate the optical flow with a deep neural network, and used another network to predict the fusion weights for merging the aligned images.
Although improved results over traditional methods~\cite{kalantari2013patch,mangiat2011spatially,kang2003high,li2016maximum} have been achieved, their method still relies on the accuracy of optical flow alignment and pixel blending, and suffers from ghosting artifacts in regions with large motion (see the second row of~\fref{fig:teaser}).
It remains a challenging problem to reconstruct ghost-free HDR videos from sequences with alternating exposures.

Recently, deformable convolution~\cite{Dai2017deformable} has been successfully applied to feature alignment in video super-resolution~\cite{wang2019edvr,Tian2020TDAN}. 
However, they are not tailored for LDR images with different exposures.
Motivated by the observation that accurate image alignment between LDR images with different exposures is difficult, and the success of deformable feature alignment for videos with constant exposure,
we introduce a two-stage coarse-to-fine framework for this problem. 
The first stage, denoted as \textit{\CoarseStage}, aligns images using optical flow in the image space and blends the aligned images to reconstruct the coarse HDR video. 
This stage can recover/remove a large part of missing details/noise from the input LDR images, but there exist some artifacts in regions with large motion.
The second stage, denoted as \textit{\RefineStage}, performs more sophisticated alignment and fusion in the feature space of the coarse HDR video using deformable convolution~\cite{Dai2017deformable} and temporal attention. 
Such a two-stage approach avoids the need of estimating highly accurate optical flow from images with different exposures, 
and therefore reduces the learning difficulty and removes ghosting artifacts in the final results.

As there is no publicly available real-world video dataset with ground-truth HDR for evaluation, comprehensive comparisons among different methods are difficult to achieve. To alleviate this problem, we create a real-world dataset containing both static and dynamic scenes as a benchmark for quantitative and qualitative evaluation. 

In summary, the key contributions of this paper are as follows:
\begin{itemize}
    \item We propose a two-stage framework, which first performs image alignment and HDR fusion in the image space and then in feature space, for HDR video reconstruction from sequences with alternating exposures.
    \item We create a real-world video dataset captured with alternating exposures as a benchmark to enable quantitative evaluation for this problem. %
    \item Our method achieves state-of-the-art results on both synthetic and real-world datasets.
\end{itemize}

\section{Related Work}
\paragraph{HDR image reconstruction}
Merging multi-exposure LDR images is the most common way to reconstruct HDR images~\cite{debevec1997recovering,Mann96hdr}.
To handle dynamic scenes, image alignment is employed to reduce the ghosting artifacts~\cite{sen2012robust,hu2013hdr,oh2014robust,ma2017robust}.
Recent methods apply deep neural networks to merge multi-exposure images~\cite{kalantari2017deep,cai2018learning,wu2018deep,yan2019attention,yan2020deep,niu2020hdr}. However, these methods rely on a fixed reference exposure (\eg, the middle exposure) and cannot be directly applied to reconstruct HDR videos from sequences with alternating exposures.
Burst denoising technique~\cite{liu2014fast,hasinoff2016burst,liba2019handheld} can also be applied to produce HDR images by denoising the \lowexposure images. 
However, this technique cannot make use of the cleaner details that exist in \highexposure images and have difficulty in handling extremely dark scenes.

There are methods for HDR reconstruction from a single LDR image. 
Traditional methods expand the dynamic range of the LDR images by applying image processing operations (\eg, function mapping and filtering)~\cite{akyuz2007hdr,banterle2009high,banterle2006inverse,banterle2008expanding,huo2014physiological,kovaleski2014high}. These methods generally cannot recover the missing details in the clipped regions. 
Recent methods proposed to adopt CNNs for single image reconstruction~\cite{eilertsen2017hdr,endo2017deep,lee2018deep,zhang2017learning,moriwaki2018hybrid,marnerides2018expandnet,liu2020single,santos2020single}. 
However, these methods focus on hallucinating the saturated regions and cannot deal with the noise in the dark regions of a \lowexposure image.

Recently, Kim~\etal~\cite{kim2019deep,kim2020jsi} proposed to tackle the problem of joint super-resolution and inverse tone-mapping.  
Instead of reconstructing the linear luminance image like previous HDR reconstruction methods, their goal was to convert a standard dynamic range (SDR) image to HDR display format (\ie, from BT.709 to BT.2020). 

\paragraph{HDR video reconstruction}
Many existing HDR video reconstruction methods rely on specialized hardware. For example, per-pixel exposure~\cite{nayar2000high}, scanline exposure/ISO~\cite{hajisharif2015adaptive,heide2014flexisp,choi2017reconstructing}, internal~\cite{tocci2011versatile,kronander2014unified} or external~\cite{mcguire2007optical} beam splitter that can split light to different sensors, modulo camera~\cite{zhao2015unbounded}, and neuromorphic camera~\cite{han2020neuromorphic}. The requirement of specialized hardware limits the widespread application of these methods.  
Recent methods also explore the problem of joint optimization of the optical encoder and CNN-based decoder for HDR imaging~\cite{metzler2020deep,sun2020learning}.

\begin{figure*} \centering
    \includegraphics[width=\textwidth]{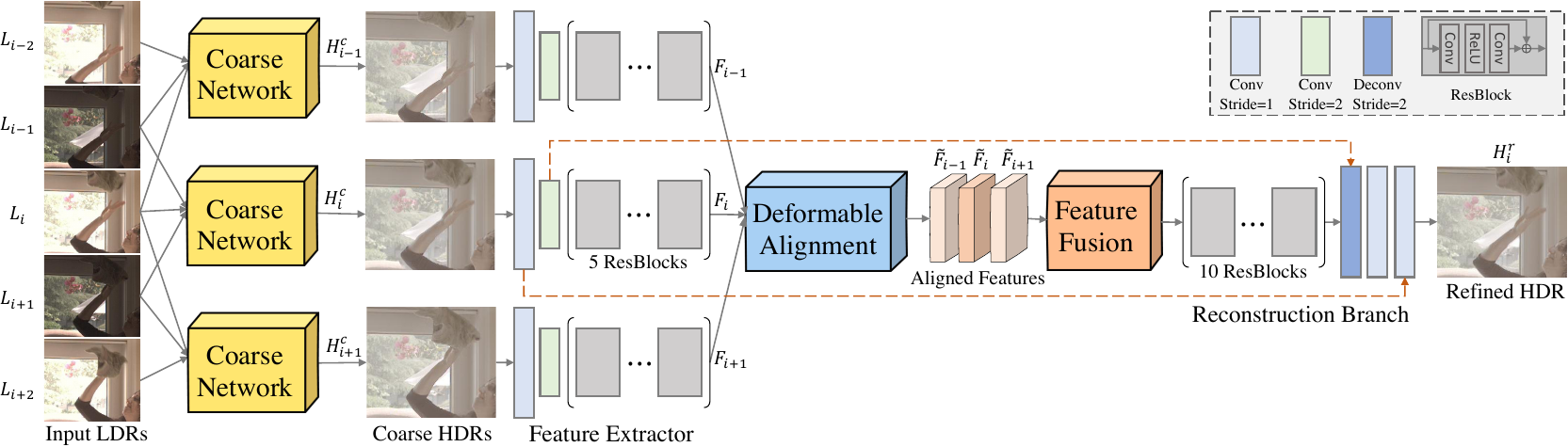}
    \caption{Network architecture of the proposed coarse-to-fine framework for videos captured with two alternating exposures.} \label{fig:network_arch}
\end{figure*}

There are works for HDR video reconstruction from sequences with alternating exposures. 
Kang~\etal~\cite{kang2003high} introduced the first algorithm of this approach by first aligning neighboring frames to the reference frame using optical flow, and then merging the aligned images to an HDR image.
Mangiat and Gibson improved this method by a block-based motion estimation and refinement stage~\cite{mangiat2010high,mangiat2011spatially}. 
Kalantari~\etal~\cite{kalantari2013patch} introduced a patch-based optimization method that synthesizes the missing exposures at each image and then reconstructs the final HDR image.
Gryaditskaya~\etal~\cite{gryaditskaya2015motion} improved~\cite{kalantari2013patch} by introducing an adaptive metering algorithm that can adjust the exposures to reduce artifacts caused by motion.
Li~\etal~\cite{li2016maximum} formulated this problem as a maximum a posteriori estimation.
Recently, Kalantari and Ramamoorthi~\cite{kalantari2019deep} introduced an end-to-end deep learning framework that contains a flow network for alignment and a weight network for pixel blending in image space. %
Different from~\cite{kalantari2019deep}, our coarse-to-fine network performs alignment and fusion sequentially in the image space and feature space for better reconstruction.

\section{The Proposed Coarse-to-fine Framework}
\subsection{Overview}
Given an input LDR video $\{\originldr_i | i = 1,\dots, n\}$ captured with alternating exposures $\{\expt_i | i = 1,\dots, n\}$\footnote{For example, the exposure can be alternated periodically in the order of  \{EV-3, EV+3, EV-3, $\dots$\} or \{EV-2, EV+0, EV+2, EV-2, $\dots$\}.}, 
our goal is to reconstruct the corresponding HDR video $\{\hdr_i | i = 1,\dots, n\}$, as shown in ~\fref{fig:teaser}.

\paragraph{Preprocessing}
Following previous methods~\cite{kalantari2013patch,li2016maximum,kalantari2019deep}, we assume the camera response function (CRF)~\cite{grossberg2003space} $\crf$ of the original input images $\originldr_i$ is known.
In practice, the CRF of a camera can be robustly estimated using a linear method~\cite{debevec1997recovering}.
As in~\cite{kalantari2019deep}, we replace the CRF of the input images with a fixed gamma curve as $\ldr_i = (\crf^{-1} (\originldr_i))^{1/\gamma}$, where $\gamma=2.2$.
This can unify input videos captured under different cameras or configurations.
Global alignment is then performed using a similarity transformation to compensate camera motions among neighboring frames. 

\paragraph{Pipeline}
Due to the existence of noise and missing details, accurate image alignment between images with different exposures is difficult.
To overcome these challenges, we introduce a two-stage framework for more accurate image alignment and fusion (see~\fref{fig:network_arch}). 
For simplicity, we illustrate our method for handling videos captured with \emph{two} alternating exposures in this paper, and describe how to extend our method for handling \emph{three} exposures in the supplementary material.

The first stage, named \emph{\CoarseStage}, aligns images using optical flow and performs HDR fusion in the image space. 
It takes three frames as input and estimates a $3$-channel HDR image for the reference (\ie, center) frame.
This stage can recover/remove a large part of the missing details/noise for the reference LDR image.
Given five consecutive LDR frames $\{\ldr_i | i = i-2, \dots, i+2\}$ with two alternating exposures, our \CoarseStage can sequentially reconstruct the coarse HDR images for the middle three frames (\ie, $\coarsehdr_{i-1}$, $\coarsehdr_{i}$, and $\coarsehdr_{i+1}$).
The second stage, named \emph{\RefineStage}, takes these three coarse HDR images as input to produce a better HDR reconstruction for the reference frame (\ie, $\refinehdr_{i}$).
It performs a more sophisticated alignment using deformable convolution and temporal fusion in the feature space.

\subsection{Coarse Reconstruction in the Image Space}
\begin{figure} \centering
    \includegraphics[width=0.48\textwidth]{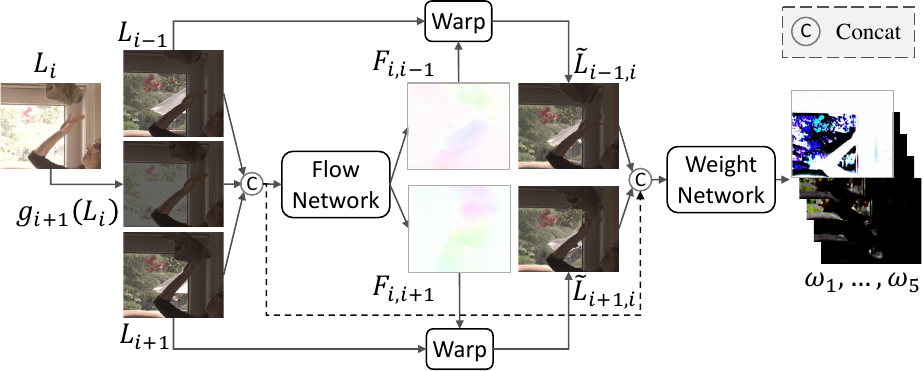}
    \caption{Overview of the \CoarseStage.} \label{fig:coarse_net}
\end{figure}

The \CoarseStage follows the design of~\cite{kalantari2019deep}, containing an optical flow estimation network, named \textit{\flownetwork}, and a blending weight estimation network, named \textit{\weightnetwork} (see \fref{fig:coarse_net}). 
The major difference is that our CoarseNet has a smaller number of feature channels, as it only performs coarse HDR reconstruction.
It first warps two neighboring frames to the center frame using optical flows, and then reconstructs the HDR image by blending the aligned images.
The network details can be found in the supplementary materials.

\paragraph{Loss function}
As HDR images are typically displayed after tonemapping, we compute the loss in the tonemapped HDR space. Following~\cite{kalantari2017deep,wu2018deep,yan2019attention,kalantari2019deep}, we adopt the differentiable $\mu$-law function:
\begin{align}
    \label{eq:mutonemap}
    \coarsetmhdr_i = \frac{\log(1+\mu \coarsehdr_i)}{\log(1+\mu)},
\end{align}
where $\coarsetmhdr_i$ is the tonemapped HDR image, and $\mu$ is a parameter controlling the compression level and is set to $5000$.
We train \CoarseStage with the L1 loss \hbox{$\coarseloss = \parallel \coarsetmhdr_i - \gttmhdr_i \parallel_1$}, 
where $\gttmhdr_i$ is the ground-truth tonemapped HDR image.
Since both the flow network and weight network are differentiable, the \CoarseStage can be trained end-to-end.

\subsection{HDR Refinement in the Feature Space}
Taking three coarse HDR images (\ie, $\coarsehdr_{i-1}$, $\coarsehdr_{i}$, and $\coarsehdr_{i+1}$) estimated by the \CoarseStage as input, the \RefineStage performs alignment and fusion in the feature space to produce better HDR reconstruction for the center frame, as the problem of missing contents or noise has been largely solved in the first stage (see the right part of~\fref{fig:network_arch}).

Our \RefineStage first extracts a $64$-channel feature for each input (\ie, $\feature_{i-1}$, $\feature_i$, and $\feature_{i+1}$) using a share-weight feature extractor. 
Features of the neighboring frames are then aligned to the center frame using a deformable alignment module~\cite{Dai2017deformable,wang2019edvr}.
The aligned features are fused using a temporal attention fusion module for the final HDR reconstruction.

\begin{figure}[t] \centering
    \includegraphics[width=0.48\textwidth]{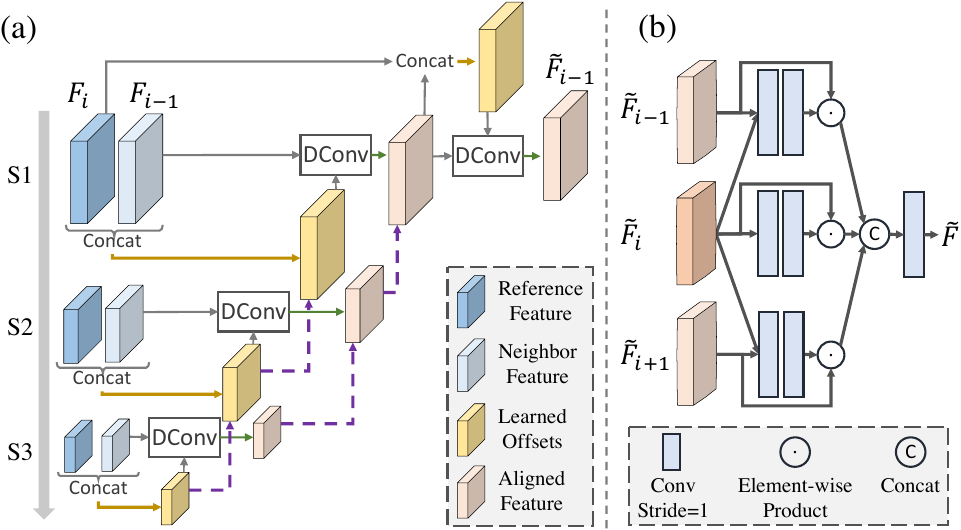}
    \caption{Structure of the (a) \featurealignmodule and (b) \featurefusionmodule.} \label{fig:PCD_fusion}
\end{figure}

\paragraph{Deformable feature alignment}
Deformable convolution~\cite{Dai2017deformable} has recently been successfully applied to feature alignment for the problem of video super-resolution (\eg, EDVR~\cite{wang2019edvr} and TDAN~\cite{Tian2020TDAN}).
The core idea of deformable feature alignment is as follows.
Given two features (\eg, $\feature_{i-1}$ and $\feature_i$) as input, 
an offset prediction module (can be general convolutional layers) predicts an offset: %
\begin{align}
    \label{eq:deconv_offset}
    \Delta p_{i-1} = f([\feature_{i-1}, \feature_{i}]).
\end{align}
With the learned offset, the neighboring feature $\feature_{i-1}$ can be sampled and aligned to the reference frame $\feature_{i}$ using deformable convolution~\cite{Dai2017deformable}: 
\begin{align}
    \label{eq:}
    \tilde{F}_{i-1} = \text{DConv}(\feature_{i-1}, \Delta p_{i-1}).
\end{align}
We adopt the pyramid, cascading and deformable (PCD) alignment module~\cite{wang2019edvr}, which performs deformable alignment in three pyramid levels, as our feature alignment module (see~\fref{fig:PCD_fusion}\,(a)).
This alignment process is implicitly learned to optimize the final HDR reconstruction.

\paragraph{Multi-feature fusion}
Given the aligned features ($\alignedfeature_{i-1}$, $\alignedfeature_{i}$, and $\alignedfeature_{i+1}$), 
we propose a temporal attention fusion module for suppressing the misaligned features and merging complementary information for more accurate HDR reconstruction (see~\fref{fig:PCD_fusion}\,(b)).
Each feature is concatenated with the reference feature as the input for two convolutional layers to estimate an attention map that has the same size as the feature. 
Each feature is then weighted by their corresponding attention map. 
Last, three attended features are concatenated and fused using a convolutional layer. %

\paragraph{HDR reconstruction}
The reconstruction branch takes the fused feature as input and regresses the HDR image ($\refinehdr_i$).
Two skip connections are added to concatenate encoder features of the \emph{reference frame} to decoder features that have the same dimensions.

\begin{figure}[t] \centering
    \raisebox{1.5\height}{\makebox[0.055\textwidth]{\makecell{\small (a) \\ \footnotesize Low Exp.}}}
    \includegraphics[width=0.175\textwidth]{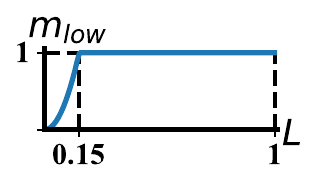} \hfill
    \raisebox{1.5\height}{\makebox[0.055\textwidth]{\makecell{\small (b) \\ \footnotesize High Exp.}}}
    \includegraphics[width=0.175\textwidth]{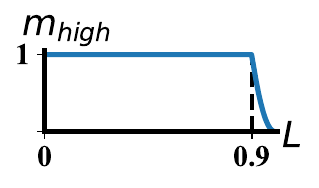} 
    \\
    \caption{Weight curves for computing the well-exposed regions for (a) low- and (b) high-exposure reference image. $L$ is the pixel value of the reference LDR image.} \label{fig:output_mask}
\end{figure}

Note that our \RefineStage aims to refine the results of \CoarseStage in the not well-exposed regions.
For a \lowexposure image, we empirically define that regions with LDR pixel values smaller than $\expolowthr$ are not well-exposed, while for a \highexposure image, regions with pixel values larger than $\expohighthr$ are not well-exposed~\cite{kalantari2013patch}.
The final predicted HDR is then computed as 
\begin{align}
    \label{eq:hdr_merge}
    \finalhdr_i = \validmask_i \odot \coarsehdr_i + (1 - \validmask_i) \odot \refinehdr_i,
\end{align}
where $\validmask_i$ is a mask indicating the well-exposed regions of the reference frame $i$, and $\odot$ is the element-wise product. \Fref{fig:output_mask} shows how $M_i$ is computed for low- and high-exposure reference image.
For example, the well-exposed mask of a \lowexposure reference image $\ldr_i$ is computed as
\begin{align}
    \label{eq:output_mask}
    \validmask_i =  \begin{cases}
        1, &\text{ if } \ldr_i >= \expolowthr \\
        (\ldr_i / \expolowthr)^2, & \text{ if } \ldr_i < \expolowthr
    \end{cases}
\end{align}

\paragraph{Loss function}
We adopt L1 loss and perceptual loss to compute the loss for \RefineStage as $\refineloss = \refineloss_{l1} + \refineloss_\text{perc}$. 
The L1 loss is defined as 
\begin{align}
    \label{eq:}
    \refineloss_{l1} =\, \parallel \finaltmhdr_i - \gttmhdr_i \parallel_1  / \parallel 1 - \validmask_i \parallel_1,
\end{align}
where $\finaltmhdr_i$ is the tonemapped image of $\finalhdr_i$. The loss is normalized by the number of not well-exposed pixels.
The perceptual loss is defined as
    $\refineloss_\text{perc} = \sum\nolimits_k \parallel \phi_k(\tmhdr_i) - \phi_k(\gttmhdr_i) \parallel_1,$
where $\phi_k(\cdot)$ extracts image features from the $k$\textsuperscript{th} layer of VGG16 network~\cite{simonyan2014very}. We use three layers \{\textit{relu1\_2}, \textit{relu2\_2}, \textit{relu3\_3}\} to compute the loss.

\section{Real-world Benchmark Dataset}
In this section, we introduce a real-world benchmark dataset for qualitative and quantitative evaluation. %

\paragraph{Existing real-world video dataset}
Currently, there is no benchmark dataset with ground-truth HDR for this problem.
The only public real-world dataset is the \kalantaridata~\cite{kalantari2013patch}, which consists of $9$ videos for dynamic scenes in RGB image format.
However, due to the lack of ground-truth HDR, previous works can only evaluate their methods qualitatively on this dataset.
In addition, this dataset is too small to be used for possible semi-supervised or unsupervised learning in the future.

\begin{table}[t]\centering
    \caption{Comparison between our dataset and the \kalantaridata~\cite{kalantari2013patch}. Frame number shows the image number. 2-Exp and 3-Exp indicate videos with two and three exposures, respectively.}
    \label{tab:dataset_comparision}
    \resizebox{0.48\textwidth}{!}{
    \large
    \begin{tabular}{*{10}{c}}
        \toprule
         &  &  \multicolumn{2}{c}{\makecell{Static Scenes \\ w/ GT}} & \multicolumn{2}{c}{\makecell{Dynamic Scenes \\ w/ GT}} & \multicolumn{2}{c}{\makecell{Dynamic Scenes \\ w/o GT}} \\
        &  &  \multicolumn{2}{c}{$6-9$ frames} & \multicolumn{2}{c}{$5-7$ frames} & \multicolumn{2}{c}{$50-200$ frames} \\
       Data & Size &  2-Exp & 3-Exp &  2-Exp & 3-Exp &  2-Exp & 3-Exp \\
        \midrule
        \cite{kalantari2013patch} & $1280\times 720$ & - & - & - & - & 5 & 4 \\
        Ours & $4096\times 2168$ & \NumOfStaticTwoExp & \NumOfStaticThreeExp & \NumOfDynamicGTTwoExp & \NumOfDynamicGTThreeExp & \NumOfDynamicTwoExp & \NumOfDynamicThreeExp \\
        \bottomrule
    \end{tabular}
    }
\end{table}

\paragraph{Dataset overview}
To facilitate a more comprehensive evaluation on real data, we captured a real-world dataset and generated reliable ground-truth HDR for evaluation. 
We used an off-the-shelf Basler acA4096-30uc camera for capturing videos with alternating exposures (\ie, two and three exposures) in a variety of scenes, including indoor, outdoor, daytime, and nighttime scenes. 
The captured videos have a frame rate of $26$ fps and a resolution of $4096\times 2168$. %

Three different types of video data are captured, namely, \staticdatalong (\staticdata), \dynamicgtdatalong (\dynamicgtdata), and \dynamicdatalong (\dynamicdata).\footnote{GT is short for the ground-truth HDR.}
\Tref{tab:dataset_comparision} compares the statistics between our dataset and \kalantaridata.

\paragraph{Static scenes with GT}
For static scenes, we captured $\NumOfStaticTwoExp$ \twoexposure and $\NumOfStaticThreeExp$ \threeexposure sequences, each with $15-20$ frames. 
The ground-truth HDR frames for static scenes were generated by merging multi-exposure images~\cite{debevec1997recovering}.
We first averaged images having the same exposure to reduce noise, and then merged multi-exposure images using a weighting function similar to~\cite{kalantari2017deep}.
For each scene, we will release $6-9$ captured frames and the generated HDR frame.

\paragraph{Dynamic scenes with GT}
Generating per-frame ground-truth HDR for dynamic videos is very challenging.
Following the strategy used for capturing dynamic HDR image~\cite{kalantari2017deep}, we propose to create image pairs consisting of input LDR frames and the HDR of the center frame.
We considered static environment and used a human subject to simulate motion in videos.

\begin{figure}[t] \centering
    \includegraphics[width=0.48\textwidth]{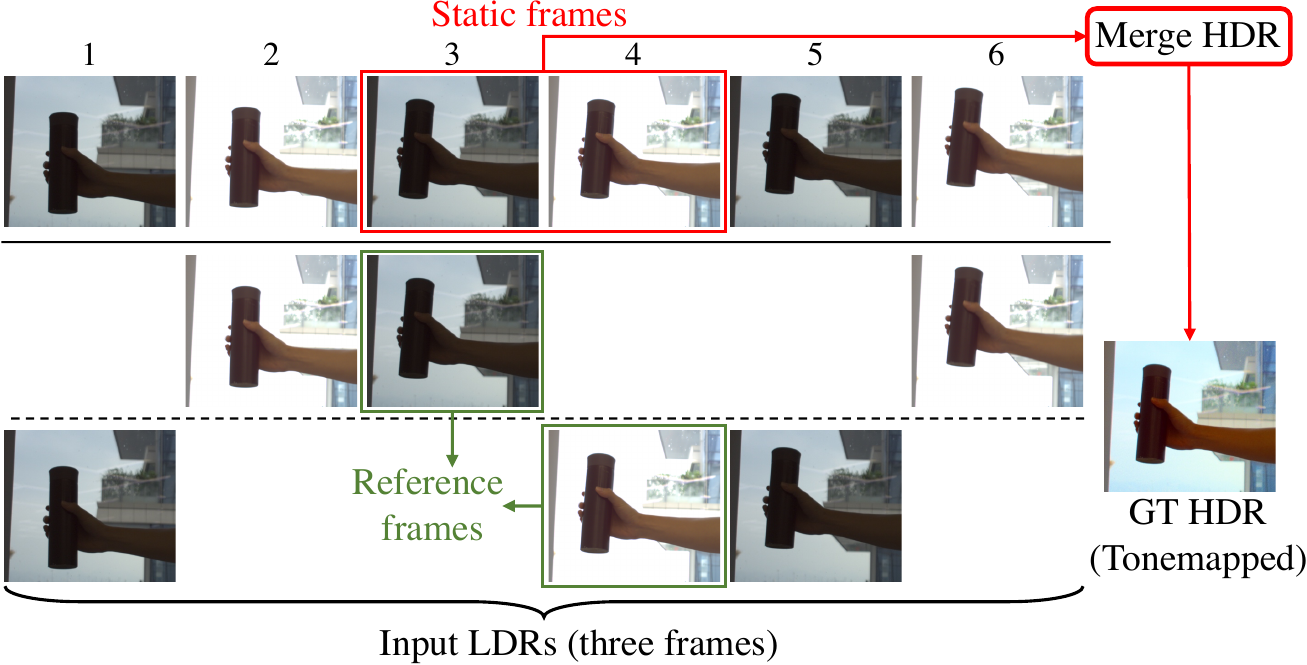}
    \caption{Illustration of generating the LDRs-HDR pairs for a \twoexposure scene (3 frames). 
    Row 1 shows the selected image sequence. Rows 2 and 3 are two sample pairs with \lowexposure and \highexposure reference frames, respectively.} \label{fig:dynamic_gt_dataset} 
\end{figure}

For each scene, we first asked the subject to stay still for $1-2$ seconds, where we can find $2$ consecutive still frames (or $3$ frames for \threeexposure) without motions for generating the HDR image for this timestamp. 
We then asked the subject to move back-and-forth (\eg, waving hands or walking). 
We selected an image sequence whose center frame was the static frame,
and arranged this sequence to be the proper LDRs-HDR pairs (see \fref{fig:dynamic_gt_dataset} for an example).
For each reference frame with GT HDR, we also created a pair with a larger motion by sampling the neighboring frames in a frame interval of $2$, which doubles the number of pairs.
In total, we created $\NumOfDynamicGTTwoExp$ and $\NumOfDynamicGTThreeExp$ pairs for the case of \twoexposure ($5$ input frames) and \threeexposure ($7$ input frames), respectively.

\begin{figure}[t] \centering
    \includegraphics[width=0.115\textwidth]{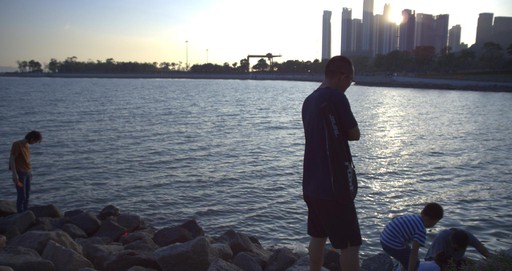}
    \includegraphics[width=0.115\textwidth]{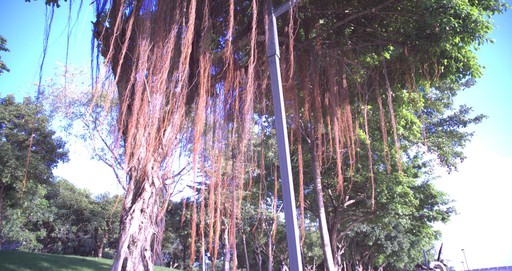}
    \includegraphics[width=0.115\textwidth]{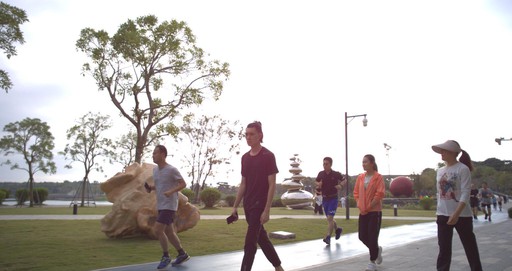}
    \includegraphics[width=0.115\textwidth]{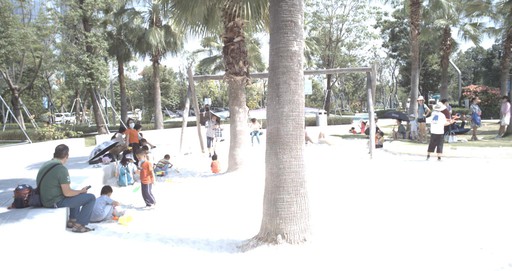}
    \\
    \includegraphics[width=0.115\textwidth]{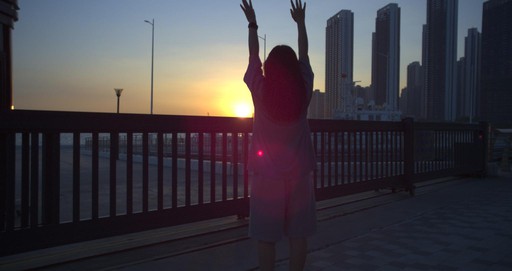}
    \includegraphics[width=0.115\textwidth]{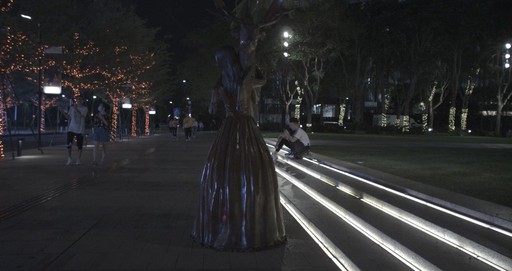}
    \includegraphics[width=0.115\textwidth]{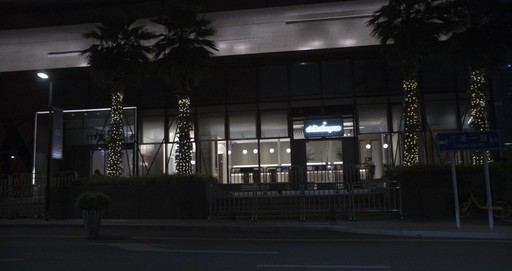}
    \includegraphics[width=0.115\textwidth]{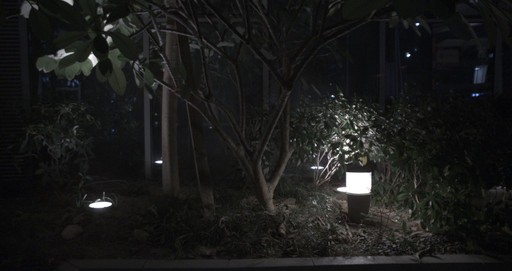}
    \\
    \caption{Sample frames in \dynamicdatalong.} \label{fig:real_dynamic_samples}
    \vspace{-1.5em}
\end{figure}

\paragraph{Dynamic scenes without GT}
We captured a larger scale dataset containing uncontrolled dynamic scenes for qualitative evaluation (see~\fref{fig:real_dynamic_samples} for examples).
Specifically, we captured $\NumOfDynamicTwoExp$ \twoexposure and $\NumOfDynamicThreeExp$ \threeexposure sequences, each contains around $100$ frames. 
This dataset can also be used for semi-supervised or unsupervised training in the future.

\paragraph{Data processing} 
We saved the raw data of the captured videos and performed demosaicing, white balancing, color correction, and gamma compression ($\gamma=2.2$) to convert the raw data to RGB data using the recorded metadata.
In this paper, we rescaled the images to $1536\times 813$ for evaluation. 
Both the captured raw data and processed images will be released.

\begin{table}[t] \centering
    \caption{Averaged results on synthetic dataset.}
    \label{tab:res_synth}
        \resizebox{0.48\textwidth}{!}{
    \begin{tabular}{l||*{3}{c}||*{3}{c}}
        \toprule
        \multicolumn{1}{c||}{} & \multicolumn{3}{c||}{\ntwoExposure} & \multicolumn{3}{c}{\nthreeExposure} \\
        Method & \PSNRT & \VDP & \VQM & \PSNRT & \VDP & \VQM \\
        \midrule
        \KalantariTOG & 37.53 & 59.07 & 84.51 & 30.36 & 56.56 & 65.90 \\
        \YanCVPR      & 39.05 & 70.61 & 71.27 & 36.28 & 65.47 & 72.20 \\
        \KalantariEG  & 37.48 & 70.67 & 84.57 & 36.27 & 65.51 & 72.58 \\
        Ours          & \B{40.34} & \B{71.79} & \B{85.71}  & \B{37.04} & \B{66.44} & \B{73.38} \\
        \bottomrule
    \end{tabular}
    }

\end{table}

\section{Experiments}
In this section, we conduct experiments on synthetic and real-world datasets to verify the effectiveness of the proposed method.
We compared our methods with \KalantariTOG, \KalantariEG, and \YanCVPR. 
\KalantariTOG is an optimization-based method and we used the publicly available code for testing.
Note that \YanCVPR is a state-of-the-art method for multi-exposure HDR image reconstruction, and we adapted it for video reconstruction by changing the network input.
We re-implemented~\cite{kalantari2019deep,yan2019attention} and trained them using the same dataset as our method. %

We evaluated the estimated HDR in terms of PSNR (in the $\mu$-law tonemapped domain), HDR-VDP-2~\cite{mantiuk2011hdr}, and HDR-VQM~\cite{narwaria2015hdr}.
HDR-VQM is designed for evaluating the quality of HDR videos.
All visual results in the experiment are tonemapped using Reinhard~\etal's method~\cite{reinhard2002photographic} following~\cite{kalantari2019deep,kalantari2013patch,kang2003high}.
In addition, a user study~\cite{bertalmio2019vision} (\ie, pair comparison test) was conducted. %

\subsection{Training Datasets and Details}
\paragraph{Synthetic training dataset}
Since there is no publicly available real video dataset with alternating exposures and their ground-truth HDR, we resort to synthetic data for training. 
Following~\cite{kalantari2019deep}, we selected $21$ HDR videos~\cite{froehlich2014creating,kronander2014unified} to synthesize the training dataset. 
Since the size of the HDR video dataset is limited,
we also adopted the high-quality Vimeo-90K dataset~\cite{xue2019video} to be the source videos.
Please refer to our supplementary material for more details.

\paragraph{Data augmentation}
As the training data was generated from clean HDR videos, the resulting input sequences lack noise in the \lowexposure images. 
To close this gap, we randomly added zero-mean Gaussian noise ($\sigma=10^{-3}$) in the linear domain of the inputs. %
We also perturbed the tone of the reference image using a gamma function ($\gamma=\exp(d), d\in [-0.7, 0.7]$) to simulate the possibly inaccurate CRF~\cite{kalantari2019deep,gil2019issues}.
Random horizontal/vertical flipping and rotation were applied. Patches of size $256\times 256$ were cropped out to be the network input.

\paragraph{Implementation details}
We trained our method using Adam optimizer~\cite{kingma2014adam} with default parameters.
We first trained the \CoarseStage with $10$ epochs using a batch size of $16$, 
and then trained the \RefineStage with $15$ epochs using a batch size of $8$.
The learning rate was initially set to $0.0001$ and halved every $5$ epochs for both networks.
We then end-to-end finetuned the whole network for $2$ epochs using a learning rate of $0.00002$.

\begin{table*}[t] \centering
    \caption{Quantitative results on the introduced real dataset. The averaged results for each exposure and all exposures are shown. \Frst{Red} text indicates the best and \Scnd{blue} text indicates the second best result, respectively.}
    \label{tab:res_real_data_wide}
        \makebox[\textwidth]{\small (a) Results on \staticdatalong (\staticdata) augmented with random global motion.} 
    \resizebox{\textwidth}{!}{
        \Large
    \begin{tabular}{l||*{2}{c}|*{2}{c}|*{3}{c}||*{2}{c}|*{2}{c}|*{2}{c}|*{3}{c}}
        \toprule
        & \multicolumn{7}{c||}{\ntwoExposure} & \multicolumn{9}{c}{\nthreeExposure} \\
        & \multicolumn{2}{c}{\LowExposure} & \multicolumn{2}{c}{\HighExposure} & \multicolumn{3}{c||}{\AllExposure} & \multicolumn{2}{c}{\LowExposure} & \multicolumn{2}{c}{\MiddleExposure} & \multicolumn{2}{c}{\HighExposure} & \multicolumn{3}{c}{\AllExposure} \\
        Method & \PSNRT & \VDP & \PSNRT & \VDP  & \PSNRT & \VDP & \VQM & \PSNRT & \VDP & \PSNRT & \VDP  & \PSNRT & \VDP & \PSNRT & \VDP & \VQM \\
        \midrule
        \KalantariTOG & \Scnd{40.00} & 73.70 & \Scnd{40.04} & \Scnd{70.08} & \Scnd{40.02} & 71.89 & \Scnd{76.22} 
                      & \Scnd{39.61} & 73.24 & \Frst{39.67} & \Frst{73.24} & \Frst{40.01} & 67.90 & \Frst{39.77} & 70.37 & 79.55 \\
        \YanCVPR      & 34.54 & 80.22 & 39.25 & 65.96 & 36.90 & 73.09 & 65.33 
                      & 36.51 & 77.78 & 37.45 & 69.79 & 39.02 & 64.57 & 37.66 & 70.71 & 70.13 \\
        \KalantariEG  & 39.79 & \Scnd{81.02} & 39.96 & 67.25 & 39.88 & \Scnd{74.13} & 73.84 
                      & 39.48 & \Scnd{78.13} & 38.43 & 70.08 & 39.60 & \Scnd{67.94} & 39.17 & \Scnd{72.05} & \Scnd{80.70} \\
        Ours          & \Frst{41.95} & \Frst{81.03} & \Frst{40.41} & \Frst{71.27} & \Frst{41.18} & \Frst{76.15} & \Frst{78.84}
                      & \Frst{40.00} & \Frst{78.66} & \Scnd{39.27} & \Scnd{73.10} & \Scnd{39.99} & \Frst{69.99} & \Scnd{39.75} & \Frst{73.92} & \Frst{82.87} \\
        \bottomrule
    \end{tabular}
    }

    \vspace{0.5em}
    \makebox[\textwidth]{\small (b) Results on \dynamicgtdatalong (\dynamicgtdata).} 
    \resizebox{\textwidth}{!}{
        \Large
    \begin{tabular}{l||*{2}{c}|*{2}{c}|*{3}{c}||*{2}{c}|*{2}{c}|*{2}{c}|*{3}{c}}
        \toprule
        & \multicolumn{7}{c||}{\ntwoExposure} & \multicolumn{9}{c}{\nthreeExposure} \\
        & \multicolumn{2}{c}{\LowExposure} & \multicolumn{2}{c}{\HighExposure} & \multicolumn{3}{c||}{\AllExposure} & \multicolumn{2}{c}{\LowExposure} & \multicolumn{2}{c}{\MiddleExposure} & \multicolumn{2}{c}{\HighExposure} & \multicolumn{3}{c}{\AllExposure} \\
        Method & \PSNRT & \VDP & \PSNRT & \VDP  & \PSNRT & \VDP & \VQM & \PSNRT & \VDP & \PSNRT & \VDP  & \PSNRT & \VDP & \PSNRT & \VDP & \VQM \\
        \midrule
        \KalantariTOG & 37.73 & 74.05 & 45.71 & 66.67 & 41.72 & 70.36 & 85.33 
                      & 37.53 & 72.03 & 36.38 & 65.37 & 34.73 & 62.24 & 36.21 & 66.55 & 84.43 \\
        \YanCVPR      & 36.41 & 85.68 & \Scnd{49.89} & \Scnd{69.90} & 43.15 & 77.79 & 78.92 
                      & 36.43 & 77.74 & 39.80 & \Scnd{67.88} & \Scnd{43.03} & \Scnd{64.74} & 39.75 & \Scnd{70.12} & 87.93 \\
        \KalantariEG  & \Scnd{39.94} & \Scnd{86.77} & 49.49 & 69.04 & \Scnd{44.72} & \Scnd{77.91} & \Scnd{87.16} 
                      & \Scnd{38.34} & \Scnd{78.04} & \Scnd{41.21} & 66.07 & 42.66 & 64.01 & \Scnd{40.74} & 69.37 & \Scnd{89.36}\\
        Ours          & \Frst{40.83} & \Frst{86.84} & \Frst{50.10} & \Frst{71.33} & \Frst{45.46} & \Frst{79.09} & \Frst{87.40} 
                      & \Frst{38.77} & \Frst{78.11} & \Frst{41.47} & \Frst{68.49} & \Frst{43.24} & \Frst{65.08} & \Frst{41.16} & \Frst{70.56} & \Frst{89.56}\\
        \bottomrule
    \end{tabular}
    }

\end{table*}

\begin{figure}[t] \centering
    \input{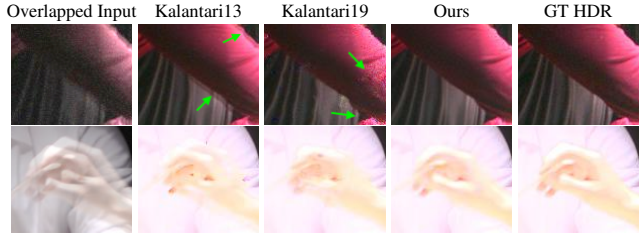}
    \caption{Visual results on the synthetic dataset.} \label{fig:res_qual_syn}
\end{figure}

\subsection{Evaluation on Synthetic Dataset}
We first evaluated our method on a synthetic dataset generated using two HDR videos (\ie, \pokerscene and \carouselscene)~\cite{froehlich2014creating}, which are not used for training. 
Each video contains $60$ frames and has a resolution of $1920\times 1080$. Random Gaussian noise was added on the \lowexposure images.
\Tref{tab:res_synth} clearly shows that our method outperforms previous methods in all metrics on the this dataset.
\Fref{fig:res_qual_syn} visualizes that our method can effectively remove the noise (top row) and ghosting artifacts (bottom row) in the reconstructed HDR.

\subsection{Evaluation on Real-world Dataset}
To validate the generalization ability of our method on real data, we then evaluated the proposed method on the introduced real-world dataset and \kalantaridata~\cite{kalantari2013patch}.

\begin{figure}[t] \centering
    \input{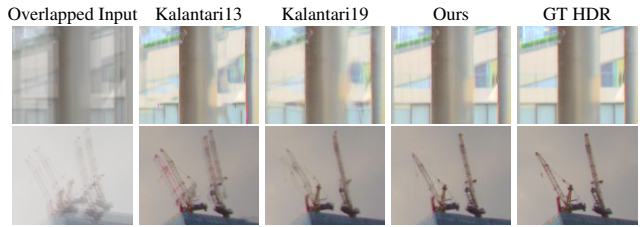}
    \caption{Visual results on \emph{static scenes} augmented with random global motion. Row~1 is for \twoexposure scene and row~2 is for \threeexposure. } \label{fig:res_qual_our_static}
\end{figure}

\begin{figure*}[t] \centering
    \input{figures/res_qual_our_dynamic}
    \caption{Visual results on \dynamicgtdatalong (\twoexposure scene).} \label{fig:res_qual_our_dynamic}
\end{figure*}

\paragraph{Evaluation on static scenes}
We evaluated our method on \staticdata augmented with random global motions (\ie, random translation for each frame in the range of $[0, 5]$ pixels).
We did not pre-align the input frames for all methods to investigate their robustness against input with inaccurate global alignment.
\Tref{tab:res_real_data_wide}\,(a) shows that our method achieves the best results for \twoexposure scenes and the most robust results for \threeexposure scenes. 
Although \KalantariTOG shows slightly better averaged PSNR values for \threeexposure scenes (\ie, $39.77$ vs. $39.75$), it suffers from the ghosting artifacts for over-exposed regions (see~\fref{fig:res_qual_our_static}).

\paragraph{Evaluation on dynamic scenes}
\Tref{tab:res_real_data_wide}\,(b) summarizes the results on \dynamicgtdata, %
where our method performs the best in all metrics. 
Compared with our method, the performance of \KalantariTOG drops quickly for dynamic scenes, as this dataset contains the more challenging local motions. 
\Fref{fig:res_qual_our_dynamic} shows that methods performing alignment and fusion in the image space~\cite{kalantari2013patch,kalantari2019deep} produce unpleasing artifacts around the motion boundaries.
In contrast, our two-stage coarse-to-fine framework enables more accurate alignment and fusion, and is therefore robust to regions with large motion and produces ghost-free reconstructions for scenes with two and three exposures.

\begin{figure*}[t] \centering
    \input{figures/res_qual_TOG13_throw}
    \caption{Visual comparison on \ThrowTowelScene scene from \kalantaridata}. \label{fig:res_qual_tog13_data} %
\end{figure*}

\paragraph{Evaluation on \kalantaridata}
We then evaluated our method on \kalantaridata.
Note that the result of \KalantariEG for this dataset is provided by the authors.
\Fref{fig:res_qual_tog13_data} compares the results for three consecutive frames from \ThrowTowelScene scene, 
where our method achieves significantly better visual results.
For a \highexposure reference frame, our method can recover the fine details of the over-exposed regions without introducing artifacts (see rows 1 and 3).
In comparison, methods based on optical flow alignment and image blending~\cite{kalantari2013patch,kalantari2019deep} suffers from artifacts for the over-exposed regions.
For a \lowexposure reference frame, compared with \KalantariTOG, our method can remove the noise and preserve the structure for the dark regions (see row 2).
Please refer to our supplementary materials for more qualitative comparisons.

\begin{wrapfigure}{r}{0.25\textwidth} \centering
    \vspace{-1.0em}
    \hspace{-1.7em}
    \includegraphics[width=0.27\textwidth]{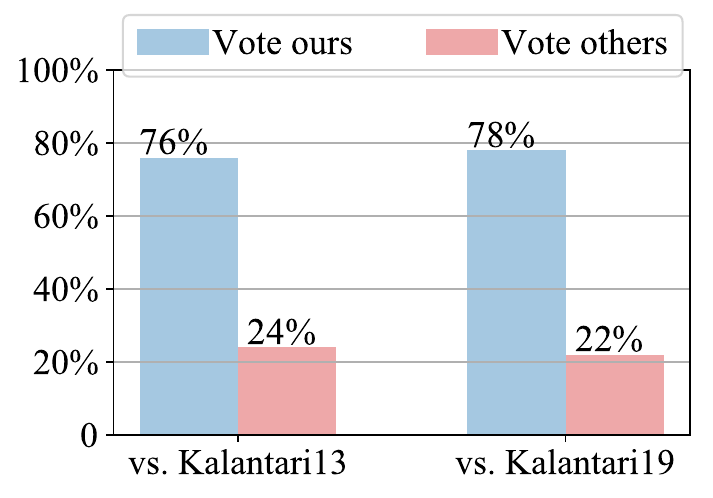}
    \caption{User study results.} \label{fig:user_study}
    \vspace{-1em}
\end{wrapfigure}

\paragraph{User study}
We also conducted a user study on the dynamic scene dataset (3-Exp) to further demonstrate the visual quality of our results (see \fref{fig:user_study}).
$33$ participants were invited to give preference on $36$ pairs of image. Note that the GT HDR was also shown for reference.
Overall, $76$\% and $78$\% of the users preferred results of our method over Kalantari13~\cite{kalantari2013patch} and Kalantari19~\cite{kalantari2019deep}, reiterating the effectiveness of our method.

\subsection{Network Analysis}
We first discussed the network parameter and runtime, and then conducted ablation study for the proposed method. 

\begin{table}[t] \centering
    \caption{Model parameter and runtime for producing an HDR frame of different resolutions.}
    \label{tab:runtime}
    \resizebox{0.48\textwidth}{!}{
        \Large
    \begin{tabular}{l*{1}{c}||*{2}{c}||*{2}{c}}
        \toprule
        \multicolumn{1}{c}{} & \multicolumn{1}{c||}{} & \multicolumn{2}{c||}{\ntwoExposure} & \multicolumn{2}{c}{\nthreeExposure} \\
        Method & \# Parameter & $1280 \times 720$ & $1920 \times 1080$ & $1280 \times 720$ & $1920 \times 1080$ \\
        \midrule
        \KalantariTOG & -      & 125s  & 185s & 300s & 520s \\
        \KalantariEG  & $9.0M$ & 0.35s & 0.59s & 0.42 & 0.64 \\
        Ours          & $6.1M$ & 0.51s & 0.97s & 0.64 & 1.09s \\
        \bottomrule
    \end{tabular}
    }
    \vspace{-1em}
\end{table}

\paragraph{Parameters and runtime}
\Tref{tab:runtime} compares the parameter and runtime of three methods. %
Note that \KalantariEG and our method were run on a NVIDIA V100 GPU, while \KalantariTOG was run on CPUs.
Our model contains $6.1$ million parameters, including $3.1M$ parameters for \CoarseStage and $3.0M$ for \RefineStage.
It takes around $1$ second for our method to produce an HDR frame with a resolution of $1920\times 1080$, which is comparable to~\KalantariEG and significantly faster than~\KalantariTOG.

\paragraph{Coarse-to-fine architecture}
To verify the design of our coarse-to-fine architecture, 
we compared our method with two baselines. 
The first one was \CoarseStage, which performs optical flow alignment and fusion in the image space (similar to \cite{kalantari2019deep}).
The second one was \SingleStageRefineStage that directly takes the LDR frames as input and performs alignment and fusion in the feature space.
Experiments with IDs 0-2 in \Tref{tab:ablation_study} show that our method achieves the best results on three datasets, demonstrating the effectiveness of our coarse-to-fine architecture.

\paragraph{Network design of the \RefineStage}
To investigate the effect of deformable alignment (DA) module and temporal attention fusion (TAF) module, 
we trained two variant models, one without DA module and one replacing TAF module with a convolution after feature concatenation. 
Experiments with IDs 2-4 in \Tref{tab:ablation_study} show that removing either component will result in decreased performance, verifying the network design of the \RefineStage.

\begin{table}[t] \centering
    \caption{Ablation study on three datasets with two alternating exposures. CNet and RNet are short for \CoarseStage and \RefineStage.} %
    \label{tab:ablation_study}
        \resizebox{0.48\textwidth}{!}{
    \Large
    \begin{tabular}{*{2}{l}||*{2}{c}||*{2}{c}||*{2}{c}}
        \toprule
        \multicolumn{1}{c}{} & \multicolumn{1}{c||}{} & \multicolumn{2}{c||}{Synthetic Dataset} & \multicolumn{2}{c||}{\staticdata} & \multicolumn{2}{c}{\dynamicgtdata} \\
        ID & Method  & \PSNRT & \VDP & \PSNRT & \VDP & \PSNRT & \VDP \\
        \midrule
        0 & CNet                    & 39.25 & 70.81 & 40.62 & 74.51 & 44.43 & 77.74 \\ %
        1 & \SingleStageRefineStage & 39.69 & 70.95 & 37.61 & 75.30 & 43.70 & 78.97 \\
        \rowcolor{gray!20}
        2 & CNet + RNet             & \B{40.34} & \B{71.79} & \B{41.18} & \B{76.15} & \B{45.46} & \B{79.09} \\
        \midrule
        3 & CNet + RNet w/o DA      & 39.72 & 71.38 & 40.52 & 74.79 & 45.09 & 78.24 \\ %
        4 & CNet + RNet w/o TAF     & 40.03 & 71.66 & 40.80 & 76.12 & 45.17 & 78.99 \\ %
        \bottomrule
    \end{tabular}
    }

    \vspace{-1em}
\end{table}

\section{Conclusion}
We have introduced a coarse-to-fine deep learning framework for HDR video reconstruction from sequences with alternating exposures.
Our method first performs coarse HDR video reconstruction in the image space and then refines the coarse predictions in the feature space to remove the ghosting artifacts.
To enable more comprehensive evaluation on real data, we created a real-world benchmark dataset for this problem.
Extensive experiments on synthetic and real datasets show that our method significantly outperforms previous methods.

Currently, our method was trained on synthetic data. Since we have captured a large-scale dynamic scene dataset, we will investigate self-supervised training or finetuning using real-world videos in the future.

\section{Acknowledgements}
This work was supported by Alibaba DAMO Academy, the Hong Kong RGC RIF grant (R5001-18), and Hong Kong RGC GRF grant (project\# 17203119).

{\small
\bibliographystyle{ieee_fullname}
\bibliography{gychen}
}

\end{document}